\documentclass{ecai} 

\usepackage{latexsym}
\usepackage{amssymb}
\usepackage{amsmath}
\usepackage{amsthm}
\usepackage{booktabs}
\usepackage{enumitem}
\usepackage{graphicx}
\usepackage{color}
\usepackage{algorithm}
\usepackage{algpseudocode}
\usepackage{float}
\usepackage{multirow}

\newtheorem{theorem}{Theorem}

\newcommand{\BibTeX}{B\kern-.05em{\sc i\kern-.025em b}\kern-.08em\TeX}

\begin{document}


\begin{frontmatter}


\paperid{598} 


\title{Learning Confidence Bounds for Classification with Imbalanced Data}


\author[A]{\fnms{Matt}~\snm{Clifford}
\thanks{Corresponding Author. Email: matt.clifford@bristol.ac.uk.}}
\author[A]{\fnms{Jonathan}~\snm{Erskine}
}
\author[A]{\fnms{Alexander}~\snm{Hepburn}
} 
\author[A]{\fnms{Raúl}~\snm{Santos-Rodríguez}
} 
\author[B]{\fnms{Dario}~\snm{Garcia-Garcia}
} 

\address[A]{University of Bristol, United Kingdom}
\address[B]{Fever Labs, United States}


\begin{abstract}
Class imbalance poses a significant challenge in classification tasks, where traditional approaches often lead to biased models and unreliable predictions. Undersampling and oversampling techniques have been commonly employed to address this issue, yet they suffer from inherent limitations stemming from their simplistic approach such as loss of information and additional biases respectively. In this paper, we propose a novel framework that leverages learning theory and concentration inequalities to overcome the shortcomings of traditional solutions. We focus on understanding the uncertainty in a class-dependent manner, as captured by confidence bounds that we directly embed into the learning process. By incorporating class-dependent estimates, our method can effectively adapt to the varying degrees of imbalance across different classes, resulting in more robust and reliable classification outcomes. We empirically show how our framework provides a promising direction for handling imbalanced data in classification tasks, offering practitioners a valuable tool for building more accurate and trustworthy models.
\end{abstract}

\end{frontmatter}




\section{Introduction}
In a supervised classification setting, class imbalance refers to scenarios where the proportion for each class in a given dataset is uneven. Real world datasets spanning many domains such as medicine, business, and bio-informatics contain some degree of imbalance \cite{Fernandez_2018_Springer}. For example, in a medical dataset depicting a rare disease diagnosis, there will inherently be more examples of healthy individuals than individuals with a positive diagnosis of the disease. In this case, the few examples of the positive diagnosis are the minority class and the healthy individuals are in the majority class. Even though there are fewer examples of the minority class, there is a high importance to correctly classify a positive instance of the disease since a delay in treatment will incur serious ramifications.

The majority of learning algorithms aim to minimise the empirical risk of the training data \cite{vapnik2013nature}, treating all mistakes as equal and not considering the different costs of errors associated with imbalanced datasets. However, without addressing imbalance during the machine learning pipeline, classifier generalisation on the minority class will be poor. This is especially true in our era of deep learning approaches which have the tendency to overfit \cite{Dablain_2023_JML, Ghosh_2022_JML} and in scenarios where the training samples do not fully represent the true data distribution.

Ideally, more samples would be gathered from the minority class to gain a better representation. However, in practice this is often not possible due to limited resources or static datasets. Popular methods of dealing with class imbalance for classification involve adjusting the training data by under and oversampling or by using a cost-sensitive training algorithm. 

Undersampling removes data points from the majority class which wastes useful information in the disregarded samples. Oversampling creates additional minority class data points. This can produce a change in the data distribution which can lead to undesirable classification behaviour. Cost-sensitive algorithms which weight minority class samples with higher importance suffer from the same effects as oversampling by replication, causing the classifier to be over confident around the minority class. More involved cost-sensitive learning algorithms are often specific to a classifier and require retraining from scratch, offering little flexibility to provide a general solution to class imbalance classification.

\begin{figure}[b]
    \centering
    \includegraphics[width=0.9\columnwidth]{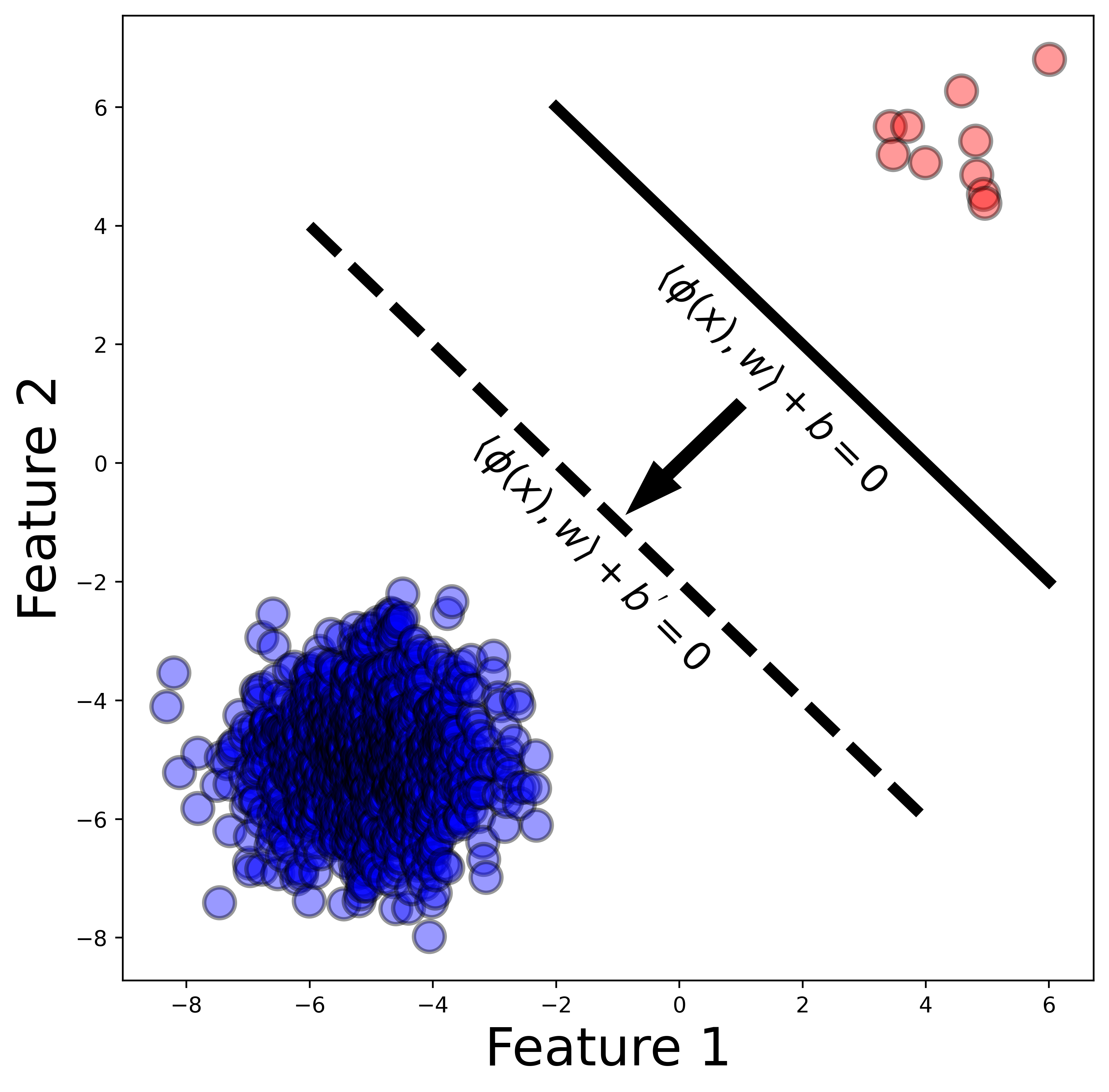}
    \caption{Training data of two normally distributed classes with class means $\{{-\boldsymbol{\mu}}, {\boldsymbol{\mu}\}}$ and where ${\boldsymbol{\mu}} = \left[{\begin{smallmatrix}5\\5\end{smallmatrix}}\right]$ and unit covariance and classifier $f(x) = sgn(\langle \phi(x), w\rangle + b)$, refer to Section \ref{sec:background} for classifier definitions. The decision boundary of a classifier (solid line) should be shifted towards the dashed line by adjusting the original bias term $b$ to $b'$ since there is a high level of uncertainty that the minority class is well defined in the training data.}
    \label{fig:moving-bias-diagram}
\end{figure}

In our work we utilise the knowledge that class imbalance exists. We embed the uncertainty associated with having fewer samples of the minority class directly into the learning process. We focus on a general class balancing method for any pre-trained binary classifier which makes decisions based upon a bias threshold, for example: Neural Networks, Support Vector Machines and logistic regression. Our method adjusts the bias term of such a pre-trained classifier, as shown in Figure \ref{fig:moving-bias-diagram}, such that the uncertainty around the distribution of the minority class is accounted for.

Adjusting the bias term is a simple solution to dealing with class imbalance. However, determining the level of adjustment is a non-trivial decision, heuristics such as a weighting of the inverse class proportions can be made but this type of approach has little guarantees around the generalisation of classification. To this end, we propose a theoretically-grounded method of adjusting the bias term by leveraging generalisation bounds provided from concentration inequalities of each class distribution.

We show that our method performs best when the pre-trained classifier has successfully learnt a good representation of the data, ideally a representation as close to separable as possible. In such cases our method outperforms or is comparable to popular techniques for balancing a pre-trained binary classifier, whilst being more theoretically-principled. In cases where the pre-trained classifier has learnt a poor representation of the data, it is advisable to retrain from scratch using a cost-sensitive training algorithm.


\section{Related Work}
\label{sec:Lit}
Learning from imbalanced data sources is an area that has attracted a lot of interest over the years \cite{Branco_2016_CSUR, He_2009_KDE, JSilva_2022_ECML}. However, most approaches to deal with class imbalance rely on addressing oversimplified versions of the problem. Nevertheless, these are popular amongst practitioners as, without intervention, standard learning algorithms lead to classifiers which perform poorly on minority classes since in order to minimise the average loss, the focus during training is skewed towards the majority class.

Methods that address class imbalance can be grouped into 3 stages of the machine learning pipeline: \emph{data adjustment} (before training), \emph{cost-sensitive learning algorithms} (during and post-training) and \emph{model adjustment} (post-training) \cite{Bahnsen_2015_ESA, Branco_2016_CSUR}. 

\paragraph{Data Adjustment}
Methods that operate before training adjust the dataset directly, with the goal of producing a dataset with balanced classes. The simplest method is to oversample the minority class by replication and/or undersample the majority class. Oversampling by replication increases the prevalence of the minority class, but suffers from changing the data distribution such that a classifier might become over confident around high density areas of replicated data points.
Random undersampling is preferable since the data distribution is altered the least \cite{Drummond_2003}, but is not suitable for datasets where the minority class contains only a few samples, as the vast amount of useful information from the majority class is discarded.

Approaches that oversample aim to reduce the amount of useful information disregarded by generating new data points within the minority class. Methods follow the patterns or properties of the data to ensure that the generated points are informative for learning and minimise the risk of the points being misleading or incorrect.

One of the most popular techniques when dealing with imbalance, combining undersampling the majority class with oversampling the minority class, is Synthetic Minority Oversampling Technique (SMOTE) \cite{Chawla_2002_JAIR}.
The oversampling procedure interpolates between existing minority class data points in order to synthesise new examples. The motivation for this is that it will create a classifier with a broader decision region for the minority class and alleviate the issues when oversampling by replication.
However, in some cases SMOTE has been shown to not respect the original data distribution, which can introduce undesirable patterns and biases to the data \cite{Elreedy_2023_JML}.

Less popular methods that use oversampling follow a similar motivation and method to SMOTE and therefore suffer from similar shortcomings as SMOTE \cite{Chawla_2003_PKDD, He_2008_IJCNN}.

Over and undersampling based methods occlude the imbalance information when transforming an unbalanced dataset into balanced. However, this knowledge of imbalance is informative, since having fewer samples in the minority class informs us that there is more uncertainty associated with the distribution of that class. The classifier should take this into account when assigning confidence around the classification of the minority class.

\paragraph{Cost-Sensitive Learning}
Class imbalance can be tackled directly during the training of a classifier. This bypasses the issues which come with adjusting the data distribution before training. 

An intuitive approach, inspired by importance sampling \cite{Kahn_1953_JORSA}, is to change the weighting or `importance' of the loss at certain data points. The simplest weighting to use is the inverse proportion a class's occurrence. These weightings are then applied in the loss function for a Neural Network (NN) \cite{Cui_2019_CVPR, Huang_2016_CVPR}, logistic regression \cite{Clifford_2023_CIKM, King_2001_PA} or Support Vector Machine (SVM) \cite{Yang_2007_IJPRAI}.

More involved cost-sensitive learning algorithms exist for SVMs \cite{Cao_2022_PAKDD}, NNs \cite{Cao_2019_NuerIPS, Cerqueira_2023_JML,raul} and Decision Trees \cite{Bahnsen_2015_ESA}. 
However, the issue with dealing with class imbalance during training is that the cost-sensitive learning algorithms are often classifier specific. Therefore, unlike the data adjustment methods which can be used with any classifier, the methods used during training do not generalise. This makes them unsuitable as a general solution to class imbalance problems.

\paragraph{Model Adjustment}
Classifiers trained on an imbalanced dataset without using a cost-sensitive algorithm still have the potential to learn useful representations of the underlying data distribution. In this case, the classifier can be adjusted post-training in order to reduce misclassification on the minority class. This is especially useful in the current era of deep learning and big data where there is a significant cost associated with training classifiers such that retraining using data balancing techniques or a cost-sensitive learning algorithm is impractical.

Post-training methods benefit from being applicable to a wide variety of classifiers, with some being classifier and training algorithm agnostic due to only acting on the outputs of a classifier. Often, the probabilities are adjusted such as changing the classification threshold by minimising the cost of misclassification using cross validation \cite{Sheng_2006_AAAI}, or using a Bayes minimum risk (BMR) decision model to quantify and minimise the risks associated with prediction \cite{Bahnsen_2013_ICMLA}. Although these methods are applicable to a wide range of classifiers, they do require a classifier which produces meaningful and well calibrated probability estimates \cite{Bahnsen_2014_SIAM}. In addition to this they require a known or estimated misclassification costs for each class.

In our work we focus on a post-training method which does not require probability outputs from a classifier, instead working directly on the bias term of a classification threshold. This permits a great deal of versatility to the range of possible scenarios class imbalance is addressed, and side steps the aforementioned issues associated with the pre and during training approaches.

\section{Background}
\label{sec:background}
In this paper we focus on classifiers that follow the generalised form $f(x) = sgn(\langle \phi(x), w\rangle + b)$, where $\phi(x)$ is a kernel or mapping function of the data $x$, $w$ represents the weights of a classifier forming $\langle \phi(x), w\rangle: \mathbb{R}^n \mapsto \mathbb{R}$ which we call the projected space of the data where the bias term $b$ thresholds the classification decision. Popular classifiers which follow this formulation include Support Vector Machines, Logistic Regression and Neural Networks.

\subsection{Training Data Bound}
In what follows, we rely on the following result concerning the symmetry property of the independent and identically distributed random variables (i.i.d.) assumption, with the full proof given in Section 5.7.1 of \cite{Blitzstein_2019_Chapman}:

\begin{theorem}
    \label{theorem:IID}
    Let $X_1,...,X_k$ be i.i.d. from a continuous distribution. Then $P(X_{a_1}<...<X_{a_k})= 1/k!$ for any permutation $a_1,...,a_k$ of $1,...,k$.
\end{theorem}

We denote $d_n = ||\phi(x_n) - \phi_S||$, where $\phi_S=\mathbb{E}[\phi(x)]$. In other words, $d_n$ is the Euclidean distance of a data point $x_n$ from its expected value in the projected space.

Given an i.i.d. training dataset $X_t = \{x_1,..., x_N\}$ there exist $N!$ possible combinations for ordering the data. From this and using Theorem \ref{theorem:IID}, the probability of a test point $x_{N+1}$ being further away from $\phi_S$ than any point in $X_t$ is
\begin{equation}
    \label{eq:iid_error}
    P\Big\{\max_{n=1,...,N} d_n < d_{N+1} \Big\} = \frac{1}{N+1}.
\end{equation}

For a more detailed intuition behind deriving Eq. \ref{eq:iid_error}, refer to Appendix \ref{appendix:training_data_bound}.

\subsection{Concentration Inequalities}
The concentration of a random variable determines how likely sets of samples will be similar, with a higher concentration resulting in a higher probability that another sample will be drawn close to its expected value. Concentration inequalities capture this by bounding the probability that a random variable will deviate from a given value. These bounds are useful to quantify the confidence on the deviation between the training set and samples drawn from the true underlying distribution after training.

In this paper, we make use of the following theorem to give a bound on the distance from the centre of mass of a set of samples to its expected value given the number of samples and the support of the distribution. The full derivation is given in Chapter 4.1 of \cite{Shawe_2004_Camb} by making use of McDiarmid's theorem \cite{McDiarmid_1989}.

\begin{theorem}[Shawe-Taylor \& Cristianini, 2004]
    \label{theorem:concentration}
    Let $S$ be an $N$ sample generated independently at random according to a distribution $P$ with support in a ball of radius $R$ in feature space, where $R=\sup_{n=1,...,N} ||\phi(x_n)||$ and $\bar{\phi}_S$ is the empirical mean of $S$. Then, with probability at least $1-\delta$ over the choice of $S$, we have
    \begin{equation}
        ||\bar{\phi}_S - \phi_S||\leq \frac{R}{\sqrt{N}} \Bigg( 2 + \sqrt{2 \ln \frac{1}{\delta}} \Bigg).
    \end{equation}
\end{theorem}

From here, we can use the triangle inequality to bound the distance from a data point to its expected value, $d_n$, in terms of the distance from the empirical mean $\bar{d}_n=||\phi(x_n) - \bar{\phi}_S||$:
\begin{equation}\label{eq:di_bounds}
    \bar{d}_n - \frac{R}{\sqrt{N}} \Bigg( 2 + \sqrt{2 \ln \frac{1}{\delta}} \Bigg) \leq d_n \leq \bar{d}_n + \frac{R}{\sqrt{N}} \Bigg( 2 + \sqrt{2 \ln \frac{1}{\delta}}\Bigg).
\end{equation}
By upper-bounding the distances from the points in the training set and lower-bounding the distance from a test point $x_{N+1}$ 
we have:
\begin{align}\label{eq:test_point_bound}
    & P\Bigg\{\max_{1\leq n \leq N} ||\phi(x_n)-\phi_S|| + \frac{2R}{\sqrt{N}} \Bigg( 2 + \sqrt{2 \ln \frac{1}{\delta}} \Bigg) \nonumber \\ 
    & <||\phi(x_{N+1}) - \phi_S|| \Bigg\} \leq \frac{1}{N+1}. 
\end{align}


\section{Method}
\begin{figure*}[ht]
    \centering
    \includegraphics[width=\textwidth]{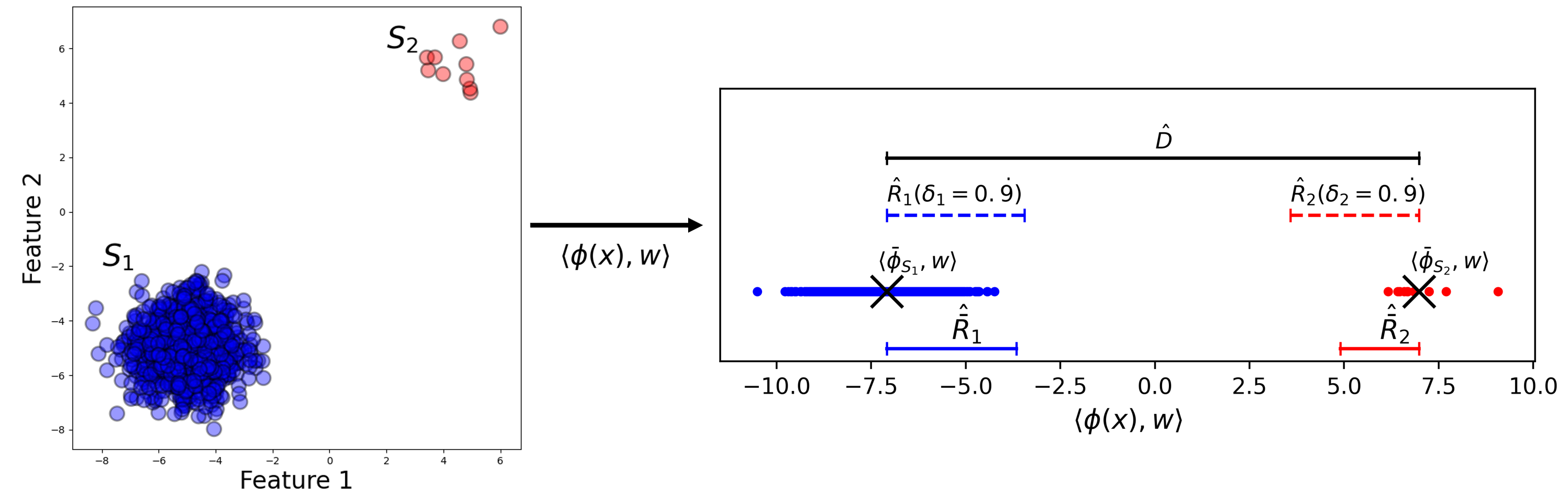}
    \caption{Left: 2D class imbalanced training dataset with samples $S_1$ and $S_2$ of each class. Right: $S_1$ and $S_2$ in the projected space from $\langle \phi(x), w\rangle$. Empirical means of each sample are shown with a cross. Distances $\hat{\bar{R}}_i$ and $\hat{D}$ are shown with solid bars. $\hat{R}_i$ are shown with dashed bars from the upper bound of Eq. \ref{R_hat_upper_bound}, where both $\delta_i = 0.\dot{9}$. }
    \label{fig:projection-diagram}
\end{figure*}

Figure \ref{fig:projection-diagram} illustrates the main concepts presented in this section. We recommended that the reader refer to it throughout this section to facilitate understanding.

Let us assume we have access to a training dataset $(x_1, y_1) , \dots , (x_n, y_n) \in \mathcal{X} \times \mathcal{Y}$ with class labels $y \in \{\pm1\}$, where samples $S_1$ and $S_2$ denote the negative and positive classes respectively. In order to separate the two classes, we can learn a classifier of the form $f(x) = sgn(\langle \phi(x), w\rangle + b)$ as described in Section \ref{sec:background}. For example, in the case of a minimum-distance classifier, the weight vector would be given by $w=\bar{\phi}_{S_1}-\bar{\phi}_{S_2}$ and $b=\frac{1}{2}||\bar{\phi}_{S_1}-\bar{\phi}_{S_2}||$. Alternately, the scoring function $f(x)$ can be learnt by commonly used frameworks such as SVMs or Neural Networks. 

In this work we focus on the role of the bias term $b$ and how to determine its value so that we can address class imbalance in a principled manner. For example, in the case of the minimum distance classifier, $b$ is set halfway between the class means. Similarly, for a vanilla SVM it is set halfway between the support vectors. Here, we introduce a method of setting $b$ in a position which is aware of class imbalance and class generalisation by utilising the concentration inequality in Eq.~\ref{eq:test_point_bound}, which remains true for the projected space $\langle \phi(x), w\rangle$ due to the linearity of the inner product.

We can interpret the classification problem in the projected space as the problem of finding the best generalising non-overlapping half-spaces in the support areas, with one half-space for each class.

Assuming the half-spaces are non-overlapping implies that the sum of the supports of each individual class can be bounded by the difference in empirical means: $\hat{\bar{R}}_1 + \hat{\bar{R}}_2 \leq \hat{D}$,
where $\hat{D} = ||\langle \bar{\phi}_{S_2}, w\rangle - \langle \bar{\phi}_{S_1}, w\rangle||$ is the distance between the projections of the classes empirical means and $\hat{\bar{R}}_i=\sup_{x\in S_i} ||\langle \phi(x), w\rangle - \langle \bar{\phi}_{S_i}, w\rangle||$ is the empirical support of class $i$.

From here, we can give an upper bound, $\hat{R}_i$, in terms of its empirical estimate $\hat{\bar{R}}_i$ as defined in Eq.~\ref{eq:di_bounds}:

\begin{equation}
    \hat{R}_i \leq \hat{\bar{R}}_i +   \frac{\hat{\bar{R}}_i}{\sqrt{N_i}} \Bigg( 2 + \sqrt{2 \ln \frac{1}{\delta_i}}\Bigg),
    \label{R_hat_upper_bound}
\end{equation}
which holds with probability $1-\delta_i$.

\subsection{Optimising the Confidence Bounds} 
Using the bounds provided, the best thing we can say about the generalisation error for a given class is that it is upper bounded by $1/(N_i+1)$ with probability $1 - \delta_i$. Our main observation is that this kind of bound suggests using the available space between $\hat{\bar{R}}_1$ and $\hat{\bar{R}}_2$ to optimise the confidence levels. Instead of using a standard approach such as setting the boundary in the middle of this space, we propose to use a loss function which averages over samples $S_1$ and $S_2$ in the projected space. We can thus propose the following loss function:

\begin{equation}\label{eq:cost_function}
    L = \sum_{i \in \{1,2\}} (1-\delta_i)\frac{1}{N_i+1} +\delta_i .
\end{equation}
 
This way, we are separating two possible sources of errors for each class: 
\begin{enumerate}
    \item The aforementioned upper bound on the generalisation error $\frac{1}{N_i+1}$ with probability $1 - \delta_i$.
    \item The cases where the bound does not hold, with the only possible upper bound being $1$ with probability $\delta_i$.
\end{enumerate}

These two sources of error provide the average error for each class via a convex combination. The confidence levels $\delta_1$ and $\delta_2$ are coupled by the fact that the half-spaces are non-overlapping. Thus, we can write:

\begin{equation}\label{eq:d_hat}
    \sum_{i \in \{1,2\}} \hat{\bar{R}}_i +  \frac{\hat{\bar{R}}_i}{\sqrt{N_i}} \Bigg( 2 + \sqrt{2 \ln \frac{1}{\delta_i}}\Bigg) = \hat{D}.
\end{equation}

This way, our problem is that of optimising Equation~\ref{eq:cost_function} s.t. the constraint in Eq.~\ref{eq:d_hat} and the additional constraints that $\delta_1, \delta_2 \in [0, 1]$ hold. We can solve for $\delta_2$ in terms of $\delta_1$, yielding:

\begin{equation}
    \delta_2 = \exp \Bigg[-0.5\bigg(\frac{B \sqrt{N_2}}{\hat{\bar{R}}_2} - 2 \bigg)^2 \Bigg],
    \label{eq:delta2_from_delta1}
\end{equation}
where 

\begin{equation*}
    B = \hat{D} - \hat{\bar{R}}_1 -  \frac{\hat{\bar{R}}_1}{\sqrt{N_1}} \Bigg( 2 + \sqrt{2 \ln \frac{1}{\delta_1}}\Bigg) - \hat{\bar{R}}_2.
\end{equation*}

Then, the confidence values $\delta_1$ and $\delta_2$ can be optimised using a constrained optimisation algorithm, using the following gradients where appropriate:
\begin{equation}
    \frac{dL}{d \delta_1} = \frac{N_1}{N_1+1} + \frac{d \delta_2}{d \delta_1} \cdot \big(\frac{N_2}{N_2+1}\big),
\end{equation}
where
\begin{align}
    \frac{d \delta_2}{d \delta_1} = - \Big(\frac{B \sqrt{N_1}}{\hat{\bar{R}}_1} - 2 \Big) \frac{\sqrt{N_2}}{\delta_1 \sqrt{N_1}} \Big(2 \ln \frac{1}{\delta_1} \Big)^{-\frac{1}{2}} \\
    \cdot \exp \Big[ -0.5(\frac{B \sqrt{N_2}}{\hat{\bar{R}}_2} - 2)^2 \Big]. \nonumber
\end{align}

Once we solve the optimisation problem for $\delta_i$, we have two values $\delta^{'}_{1}$ and $\delta^{'}_{2}$, and the new boundary is set with the bias term at the point where the two half-spaces meet which is defined as
\begin{equation}
    \label{eq:new_bias}
    b' = \langle \bar{\phi}_{S_1}, w\rangle + \hat{\bar{R}}_1 + \frac{\hat{\bar{R}}_1}{\sqrt{N_1}} \Bigg( 2 + \sqrt{2 \ln \frac{1}{\delta^{'}_{1}}}\Bigg).
\end{equation}

\subsection{Relaxing the Constraints}
\label{sec:slacks}
The method proposed so far assumes linearly separable data in the projected space $\phi(x)$, with sufficient distance between the half-spaces such that Equation \ref{eq:d_hat} can be satisfied. In practice this assumption will usually not hold. Therefore, we propose to consider slack variables into the optimisation process. 

Inspired by soft-margin SVMs \cite{cortes1995support}, let us introduce a binary slack variable, $\xi_n$, per instance in the training set, where $\xi_n \in \{0, 1\}$ defined such that $\xi_n = 0$ for $x_i$ that contribute towards $\hat{\bar{R}}_i$ and $N_i$, and $\xi_n = 1$ for $x_i$ which are allowed to exist outside of the supports of the data and do not contribute towards $\hat{\bar{R}}_i$ and $N_i$. 

The loss function is adjusted by proportionally penalising the number of points outside of the supports from each class, changing Equation \ref{eq:cost_function} to 
\begin{equation}
\label{eq:cost_slacks}
    L_{sv} = L + \alpha \sum_{n \in \{1,...,N\}} \xi_n,
\end{equation}
where $\alpha$ is the penalisation factor.

The optimisation of $L_{sv}$ is less straightforward than $L$. We iteratively set $\xi_n = 1$ for class support instances, as these are the points directly contributing to $\hat{\bar{R}}_i$. Instances from each class are selected such that $\sum_{n \in \{1,...,N\}} \xi_n$ for each class matches the class imbalance present in the training set. The final method is described in pseudocode in Algorithm \ref{alg:ours}. 
We provide an alternative approach using continuous slack variables in Appendix \ref{appendix:cont_slacks}.

\begin{algorithm}
\caption{Our Method}
\label{alg:ours}
\begin{algorithmic}[1]
    \Require Given an optimisation budget $k$ ($k \leq N$) and a trained classifier $f(x) = sgn(\langle \phi(x), w\rangle + b)$
    \For{$m=[0,...,k]$}
    \For{$i=[0, 1]$}
    \State $\sum \xi_n \gets m$ according to Sec. \ref{sec:slacks}
    \State compute $\langle \bar{\phi}_{S_i}, w\rangle$, $\hat{\bar{R}}_i$ and $N_i$
    \EndFor
    \State Optimise Eq. \ref{eq:cost_slacks} subject to the constraint in Eq. \ref{eq:d_hat}
    \State Store the loss value found $L_{sv}$
    \EndFor
    \State Pick $\delta^{'}_{1}$ corresponding to the lowest loss value
    \State Use the optimised $\delta^{'}_{1}$ in Eq. \ref{eq:new_bias} to obtain $b^{'}$
    \State Change classification boundary to $sgn(\langle \phi(x), w\rangle + b^{'})$
\end{algorithmic}
\end{algorithm}

\section{Experiments}
In this section we evaluate our method against popular class balancing techniques from the literature. First, we show an illustrative example on a simple synthetic dataset so that the reader can solidify their understanding of our method. After this, we showcase our method on commonplace datasets used for bench-marking class imbalance methods.

\paragraph{Model} We train a classifier using a standard learning algorithm on imbalanced data as our baseline. This baseline classifier is used as the pre-trained classifier $f(x)$ for our method and the two model adjustment methods presented below. We showcase our method on three baseline classifiers: Logistic Regression (LR), SVM and a fully connected Neural Network (NN).

\paragraph{Baselines} We compare our method against widely adopted methods from each stage of the machine learning pipeline presented in Section \ref{sec:Lit}. For data adjustment (before training) we use SMOTE \cite{Chawla_2002_JAIR}, for model adjustment (post-training) we use both Thresholding \cite{Sheng_2006_AAAI} and Bayes Minimum Risk (BMR) \cite{Bahnsen_2013_ICMLA, Bahnsen_2014_SIAM}. Thresholding follows the suggested method of assigning the class costs inversely proportional to the imbalance \cite{Sheng_2006_AAAI}. Most cost-sensitive learning algorithms (during training) are classifier specific, so we use the simplest method from this category by giving more importance to the minority class during training, named `Balanced Weights' (BW) from here onwards. This is achieved by the `balanced' class weights parameter provided by scikit-learn for LR and SVM \cite{scikit-learn} and directly applying it within the loss function for each sample for the NN. The weights associated with each sample are inversely proportional to their class's occurrence in the training data.

\paragraph{Evaluation Metrics}
Conventional evaluation metrics such as accuracy are not well suited to evaluating class imbalance datasets since they treat all errors equally. Metrics which rely on the precision-recall curve such as the F1 score or the G-mean are better suited to class imbalance datasets \cite{Branco_2016_CSUR, Fernandez_2018_Springer, He_2008_IJCNN}. 

In all of our experiments we follow the convention of labeling the minority class as positive and the majority class as negative so that focus is given to the performance on the minority class \cite{Chawla_2002_JAIR, Kubat_1997_ICML}.

\paragraph{Implementation Details}
Our experiments make use of the imbalance-learn package \cite{Lemaitre_2017_JMLR}, costcla \cite{Bahnsen_costcla}, scikit-learn \cite{scikit-learn}. 
Unless specified otherwise, all experiments use the default parameters from each package.

Since we use datasets with a small number of minority class points, the train/test splits can significantly skew the results. To gain unbiased results, we take the mean and standard deviation of 10 randomly sampled train and test sets for the evaluation of each dataset.

All of the experiments can be reproduced using our open source python package \cite{clifford2024_code}.
Our package provides practitioners with the ability to apply our class balancing method to their own data and classifiers with minimal code adaptation from a standard machine learning pipeline. 
All experiments in this paper are run on an AMD Ryzen 9 5950X CPU. 

For our method $\alpha=1$ and we optimise Equation \ref{eq:cost_slacks} using the gradient based constrained optimisation algorithm Sequential Least Squares Programming (SLSQP) \cite{kraft1988software} with the default parameters provided by SciPy \cite{SciPy_2020}.

\subsection{Synthetic Example}
\label{sec:synthetic}

\paragraph{Data}
We create a linearly separable 2D synthetic dataset with each class following a Gaussian distribution, namely, ${\mathcal {N}}({\boldsymbol {\mu }},\,{\boldsymbol {\Sigma }})$, with class means 
$\{{-\boldsymbol{\mu}}, {\boldsymbol{\mu}\}}$ and identity covariance,
where ${\boldsymbol{\mu}} = \left[{\begin{smallmatrix}1\\1\end{smallmatrix}}\right]$.
To illustrate our method, we choose a highly class imbalanced sampling for the training data, where $N_1=1000$ and $N_2=10$. The number of test data points is $1000$ for each class to give a true representation of each class distribution.

\begin{figure}[b]
    \centering
    \includegraphics[width=1\columnwidth]{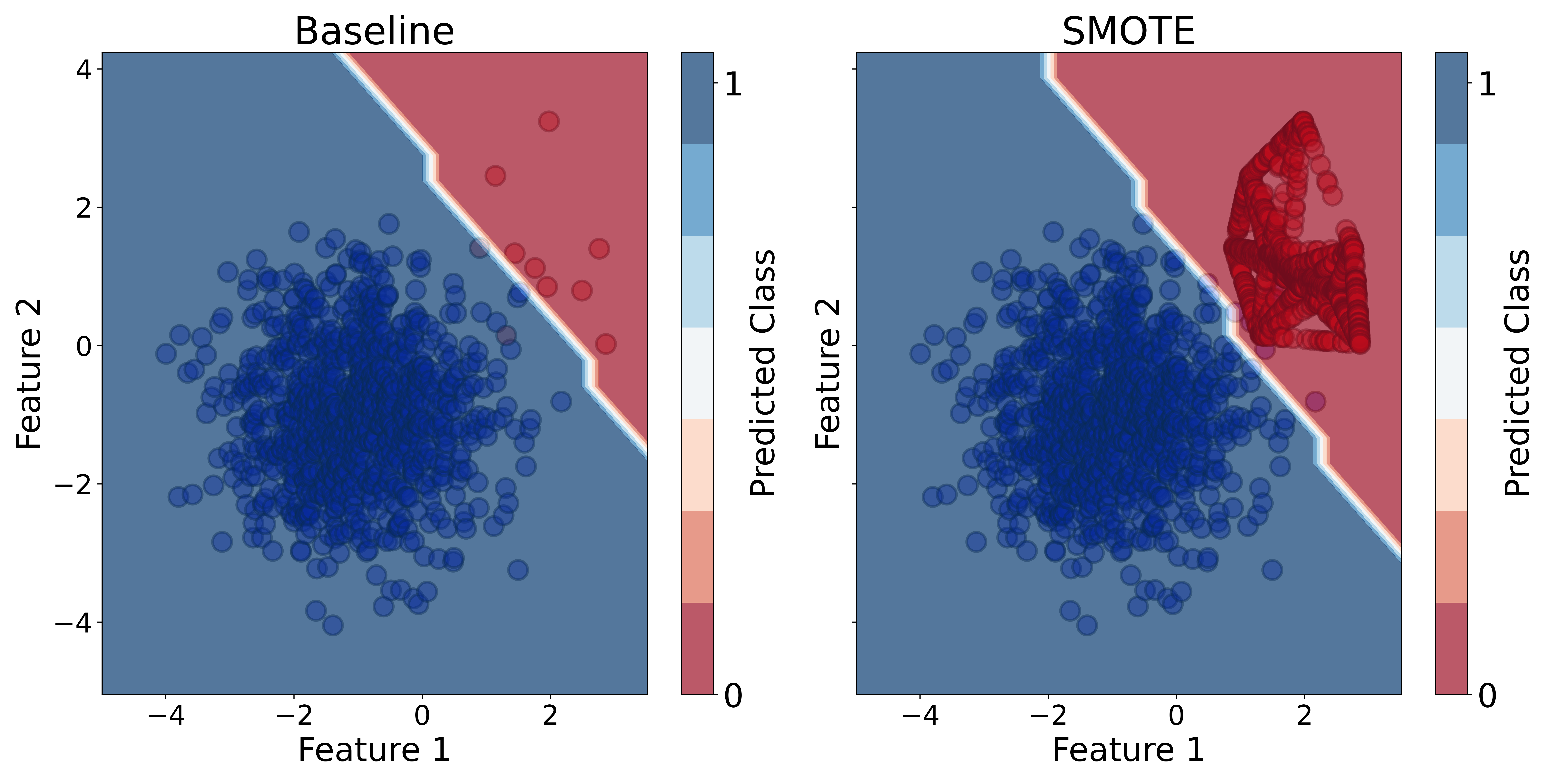}
    \caption{Logistic Regression trained on synthetic data described in Section~\ref{sec:synthetic}. Training data points are blue and red circles for the majority and minority classes respectively. Decision boundary is shown with a white line.}
    \label{fig:sythetic-LR-clfs}
\end{figure}

\paragraph{Classifier}
We train a Logistic Regression classifier~\cite{scikit-learn}. We use a standard (not cost-sensitive) learning algorithm on the original data and using SMOTE \cite{Chawla_2002_JAIR} adjusted training data. Figure \ref{fig:sythetic-LR-clfs} shows the training data and decision surface for the baseline classifier (trained on the original data) and SMOTE. We additionally train in a cost sensitive manner using balanced weights (BW).

\begin{figure}[b]
    \centering
    \includegraphics[width=\columnwidth]{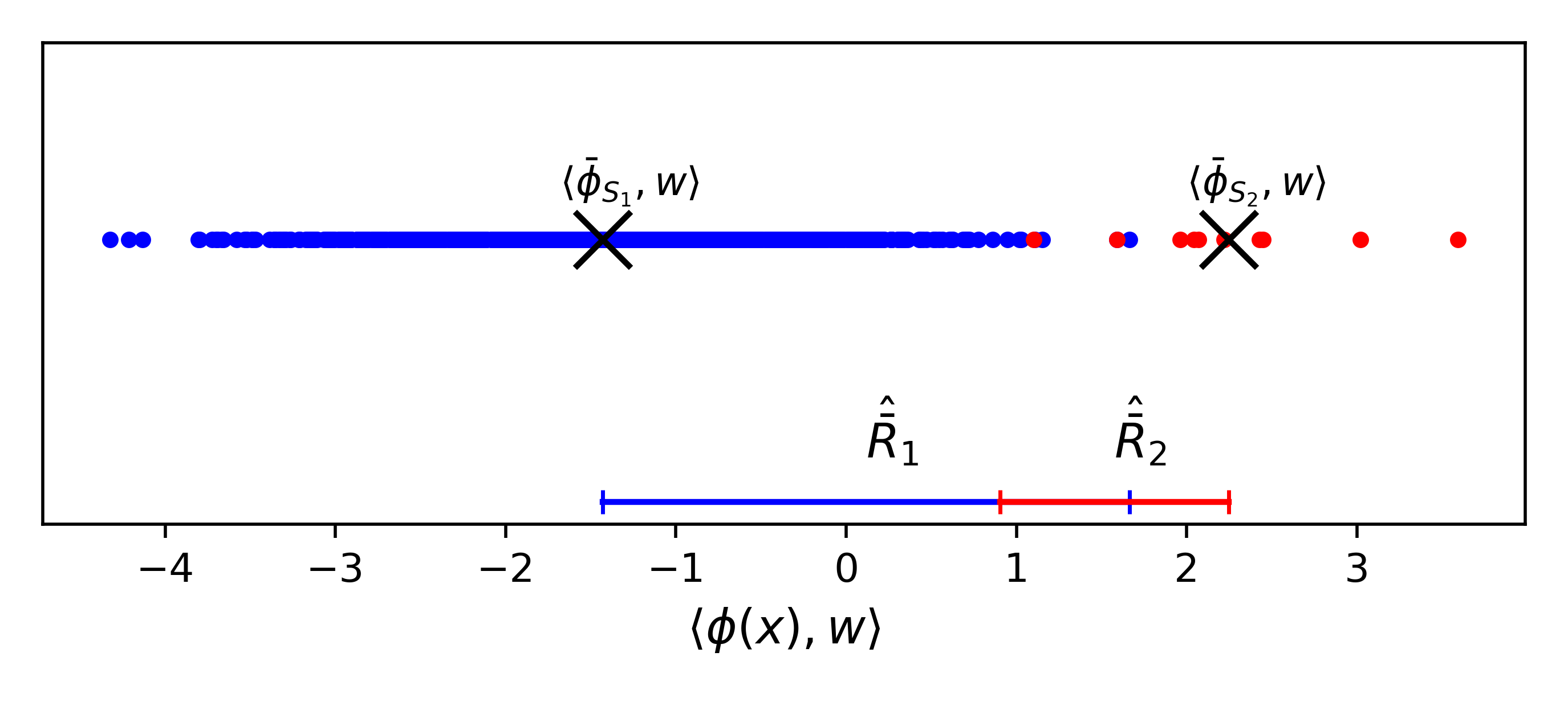}
    \caption{Projected space from Logistic Regression of the training data. Training data points are blue and red circles for the majority and minority classes respectively. $\hat{\bar{R}}_1$ and $\hat{\bar{R}}_2$ are overlapping since the classifier has not been able to linearly separate the training data in the projected space.}
    \label{fig:sythetic-deltas-data}
\end{figure}

After training, we adjust the decision boundary using Thresholding~\cite{Sheng_2006_AAAI} and Bayes Minimum Risk (BMR)~\cite{Bahnsen_2014_SIAM} from the literature as well as using our proposed method. Figure~\ref{fig:sythetic-deltas-data} shows the projected space of the training data from the classifier trained on the original (non-SMOTE) dataset.

\begin{figure}[b]
    \centering
    \includegraphics[width=\columnwidth]{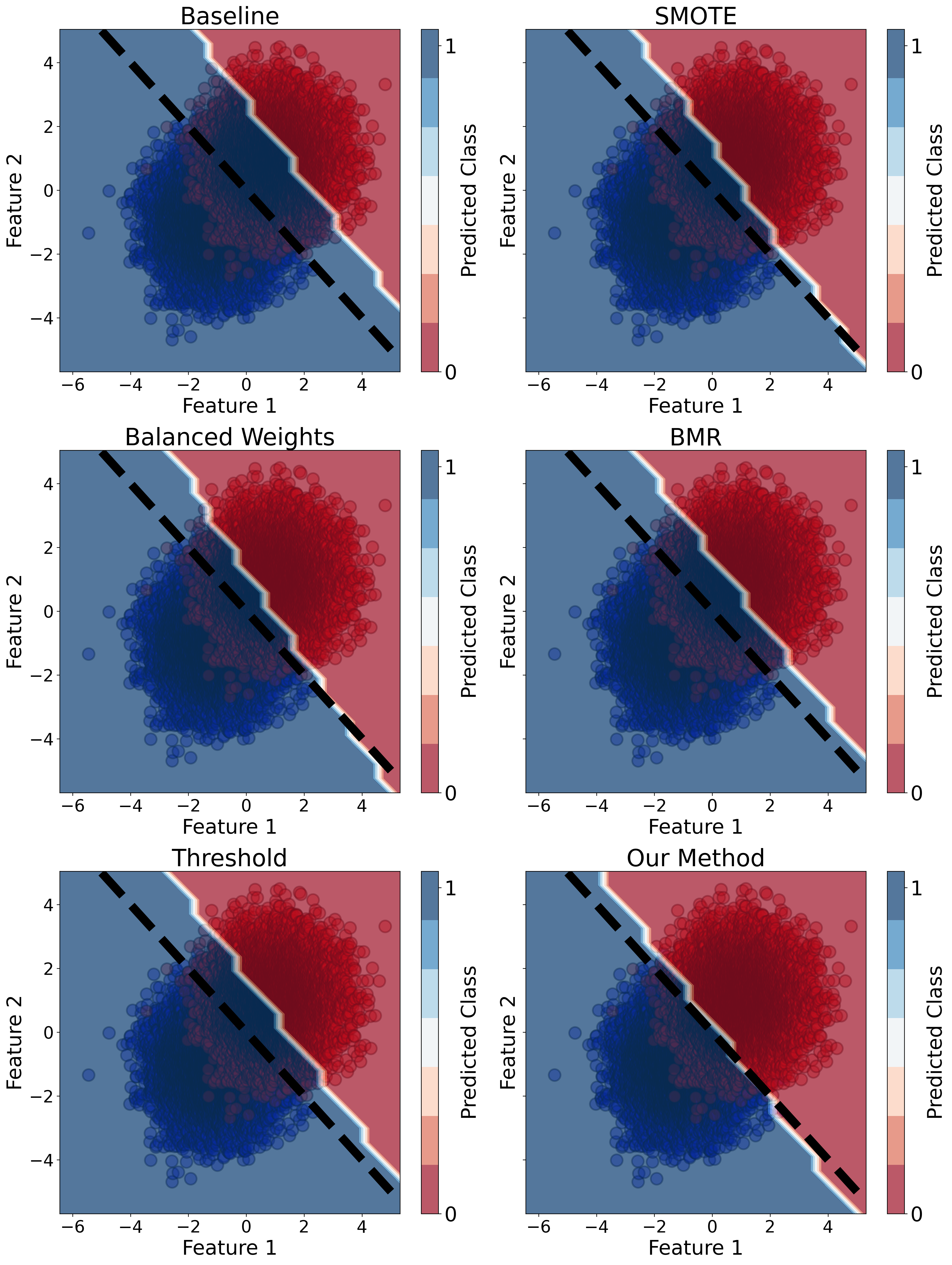}
    \caption{Comparison of methods on the test dataset from Section \ref{sec:synthetic}. The Bayes optimal classifier is shown with a dashed line. Successful class balancing methods push the boundary away from the minority class (red).}
    \label{fig:sythetic-test}
\end{figure}

\paragraph{Results}
We can see in Figure~\ref{fig:sythetic-LR-clfs} that the classifier trained on the original data prioritises making fewer mistakes on the majority class (blue), leading to poor generalisation on the minority class (top left Figure~\ref{fig:sythetic-test}). 
SMOTE balances the classes by sampling within the existing points in feature space, and although this leads to balanced classes, the data distribution of the minority class has a significantly reduced coverage of data points compared to the real distribution. This concentrated minority class introduces an incorrect bias to the data which leads to suboptimal generalisation on the minority class.

\begin{figure}[b]
    \centering
    \includegraphics[width=\columnwidth]{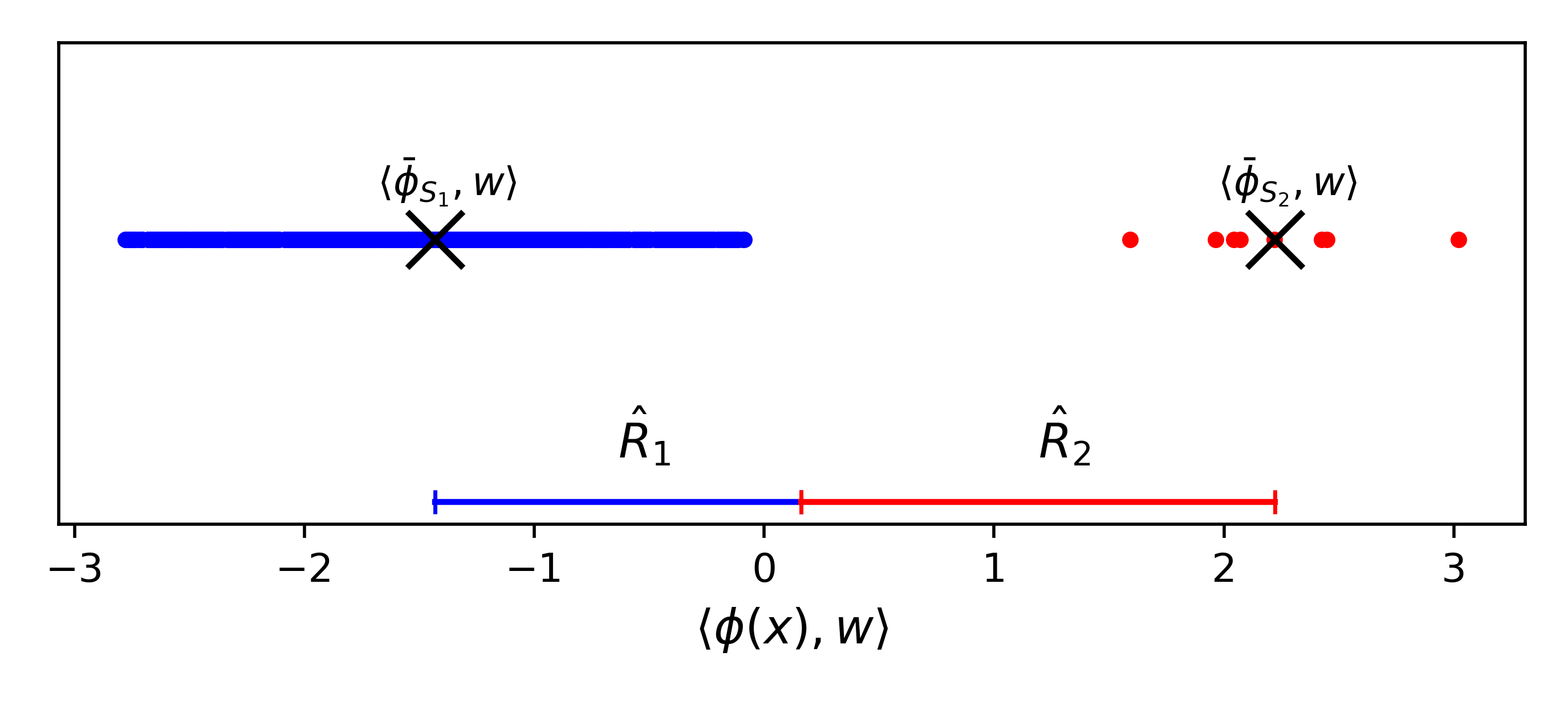}
    \caption{Projected space from Logistic Regression of the training data. Training data points where $\xi_n=0$ after optimising $\delta_i$ are shown with blue and red circles for the majority and minority classes respectively. Where $\hat{\bar{R}}_1$ and $\hat{\bar{R}}_2$ meeting from the optimised $\delta_i$ is given as the new decision boundary for classification.}
    \label{fig:sythetic-deltas-optimised}
\end{figure}

Figure \ref{fig:sythetic-deltas-optimised} shows the projected space after a minimum of Equation \ref{eq:cost_slacks} is found. Finding the solution is relatively fast, taking 3 seconds on our 16 core CPU. We can see that $\hat{R}_2$ is bigger than $\hat{R}_1$, this is due to the higher levels of uncertainty associated with the few samples present in the minority class. This extra `space' given to the minority class gives a level of caution needed for good generalisation when the minority class is scarce.

Figure \ref{fig:sythetic-test} shows the test dataset and decision surface of our method against the baseline, Balanced Weights (BW), SMOTE and threshold adjusted classifiers. If we compare the decision surfaces to the evaluation in Table \ref{tab:synethic-eval}, we can see that the class balancing methods which push the boundary the furthest away from the minority class (red) perform the best. The Baseline performs the worst overall as the classifier overfits to the minority class, leaving little room for generalisation, we also see that SMOTE and BW remain over-confident on the small area of the minority class within the training dataset.

Empirically in Table \ref{tab:synethic-eval}, our method outperforms all others across all the metrics evaluated on the test set ($\{N_1, N_2\}=1000$), with the two probability adjustment methods (BMR and Threshold) having a similar performance.

\begin{table}[h]
\caption{Evaluation on the synthetic test dataset. The best performing method for each evaluation metric is shown in bold. The minority class is treated as the positive label for all metrics. Note that the accuracy metric is balanced since the test dataset as equal class proportions.}
\begin{center}
\label{tab:synethic-eval}

\begin{tabular}{@{}lccc@{}}\toprule
Methods & Accuracy & G-Mean & F1 \\
\midrule
Baseline & $.667 \pm .02$ & $.577 \pm .04$ & $.500 \pm .05$ \\
SMOTE \cite{Chawla_2002_JAIR} & $.896 \pm .03$ & $.892 \pm .03$ & $.888 \pm .04$ \\
BW & $.902 \pm .02$ & $.899 \pm .02$ & $.896 \pm .03$ \\
BMR \cite{Bahnsen_2014_SIAM} & $.892 \pm .04$ & $.887 \pm .05$ & $.882 \pm .06$ \\
Thresh \cite{Sheng_2006_AAAI} & $.892 \pm .04$ & $.887 \pm .05$ & $.882 \pm .06$ \\
Our Method & $\textbf{.911} \pm .01$ & $\textbf{.909} \pm .01$ & $\textbf{.914} \pm .01$ \\
\bottomrule
\end{tabular}

\end{center}
\end{table}

\subsection{Class Imbalanced Datasets}
\paragraph{Data}
We evaluate our method on medical datasets, as most clinical conditions analysed in these affect small portions of the overall population, making them particularly interesting. Here, we choose those frequently used in the class imbalanced literature \cite{Chawla_2002_JAIR, Sheng_2006_AAAI}. For Pima Diabetes and Breast Cancer datasets we randomly sample a training set which contains a higher level of class imbalance (10:1) to fully showcase our method's capabilities, this comes with the benefit of providing enough minority class samples to create a roughly balanced test set.

Table \ref{tab:datasets} shows the details of the datasets. We scale all dataset features to $[-1,1]$. For the Hepatitis dataset we remove the features `Protime', `Alkphosphate' and `Albumin' and 18 instances due to missing values. For the Heart disease dataset, we remove the features `ca' and `thal', as they contain missing values. To create an unbalanced dataset, we combine the heart disease types 3 and 4 as the positive class. For the MIMIC-III dataset~\cite{Johnson_2016_mimic}, we perform ICU negative outcome prediction, using the data pre-processing outlined in~\cite{McWilliams_2019_BMJ_UCI}.

\begin{table}[h]
\caption{Datasets used for evaluation. Negative (majority) and positive (minority) class imbalance is shown for the random train and test data splits.}
\begin{center}
\label{tab:datasets}
\begin{tabular}{@{}lrrrr@{}}
\toprule
Dataset & Attributes & Instances & Train (N/P) & Test (N/P) \\ \midrule
Pima Diabetes & $8$ & $768$ & $250/25$ & $250/243$ \\
Breast Cancer & $30$ & $569$ & $178/17$ & $179/195$ \\
Hepatitis & $16$ & $137$ & $55/13$ & $56/13$ \\
Heart Disease & $11$ & $212$ & $82/24$ & $82/24$ \\
MIMIC ICU & $19$ & $7420$ & $3317/392$ & $3318/393$ \\ \bottomrule
\end{tabular}
\end{center}
\end{table}

\paragraph{Classifiers}
For the Pima Diabetes, Breast Cancer, Hepatitus and Heart Disease datasets we train an SVM classifier with Radial Basis Function (RBF) kernel. The SVM has hyper-parameters: `C', the squared L2 penalty for regularisation and `gamma', the RBF kernel coefficient. To optimise these hyper-parameters, we perform a grid search using five fold cross validation. For the MIMIC ICU dataset we train a fully connected Neural Network (MLP) with hidden layer sizes $(100, 500, 1000, 500, 100)$.

\paragraph{Results}
Table \ref{tab:all_datasets} show the evaluation results on the datasets from Tab.~\ref{tab:datasets}. Our method performs well for the balanced metrics (G-Mean and F1), where it outperforms the other methods, showcasing its effectiveness in class imbalance datasets with few minority samples. This, however, is at the expense of worse accuracy, where some examples from the majority class become misclassified. The exception to this is for the Pima Diabetes and Breast Cancer datasets where the accuracy is from a roughly balanced test set and is therefore increased.

It is worth noting that SMOTE performs poorly on the Breast Cancer and Hepatitis and Heart Disease datasets, resulting in worse scores than the Baseline across all metrics. In these cases, the few samples from the minority class do not provide enough information in feature space for SMOTE to effectively be able to synthesise meaningful data points and instead introduces additional biases for the classifier to overfit to.

There are other cases where balancing techniques are detrimental. For the Heart Disease dataset, the Balanced Weights (BW) method performs worse than the Baseline for all metrics. For the MIMIC ICU dataset, the BW method has a decrease in F1 compared to the Baseline and the BMR method increases the majority class predictive performance, as there is an increase in accuracy but a decrease in G-Mean and F1 compared to the Baseline.

\begin{table}[h]
\caption{Evaluation of all medical datasets.}
\begin{center}
\label{tab:all_datasets}

\begin{tabular}{@{}llccc@{}}
\toprule
& Methods & Accuracy & G-Mean & F1 \\
\midrule
\multirow{6}{*}{\rotatebox{90}{Pima Diabetes}}
& Baseline & $.540 \pm .04$ & $.181 \pm .23$ & $.137 \pm .18$ \\
& SMOTE \cite{Chawla_2002_JAIR} & $.595 \pm .01$ & $.495 \pm .03$ & $.396 \pm .04$ \\
& BW & $.585 \pm .02$ & $.476 \pm .06$ & $.373 \pm .08$ \\
& BMR \cite{Bahnsen_2014_SIAM} & $.661 \pm .02$ & $.628 \pm .04$ & $.575 \pm .05$ \\
& Thresh \cite{Sheng_2006_AAAI} & $.661 \pm .02$ & $.628 \pm .04$ & $.575 \pm .05$ \\
& Our Method & $\textbf{.672} \pm .03$ & $\textbf{.651} \pm .05$ & $\textbf{.611} \pm .08$ \\
\midrule
\multirow{6}{*}{\rotatebox{90}{Breast Cancer}}
& Baseline & $.917 \pm .02$ & $.917 \pm .02$ & $.913 \pm .02$ \\
& SMOTE \cite{Chawla_2002_JAIR} & $.793 \pm .09$ & $.768 \pm .12$ & $.737 \pm .14$ \\
& BW & $.921 \pm .02$ & $.921 \pm .02$ & $.918 \pm .02$ \\
& BMR \cite{Bahnsen_2014_SIAM} & $.887 \pm .08$ & $.880 \pm .09$ & $.869 \pm .11$ \\
& Thresh \cite{Sheng_2006_AAAI} & $.935 \pm .06$ & $.932 \pm .07$ & $.929 \pm .09$ \\
& Our Method & $\textbf{.943} \pm .02$ & $\textbf{.942} \pm .02$ & $\textbf{.946} \pm .02$ \\
\midrule
\multirow{6}{*}{\rotatebox{90}{Hepatitis}}
& Baseline & $\textbf{.827} \pm .06$ & $.623 \pm .15$ & $.485 \pm .16$ \\
& SMOTE \cite{Chawla_2002_JAIR} & $.804 \pm .02$ & $.299 \pm .31$ & $.224 \pm .24$ \\
& BW & $.768 \pm .09$ & $.681 \pm .09$ & $.499 \pm .10$ \\
& BMR \cite{Bahnsen_2014_SIAM} & $.804 \pm .05$ & $.656 \pm .11$ & $.493 \pm .10$ \\
& Thresh \cite{Sheng_2006_AAAI} & $.786 \pm .08$ & $.686 \pm .09$ & $.516 \pm .09$ \\
& Our Method & $.775 \pm .09$ & $\textbf{.714} \pm .05$ & $\textbf{.535} \pm .08$ \\
\midrule
\multirow{6}{*}{\rotatebox{90}{Heart Disease}}
& Baseline & $\textbf{.885} \pm .02$ & $.784 \pm .04$ & $.718 \pm .05$ \\
& SMOTE \cite{Chawla_2002_JAIR} & $.844 \pm .03$ & $.700 \pm .07$ & $.602 \pm .09$ \\
& BW & $.842 \pm .02$ & $.781 \pm .06$ & $.664 \pm .06$ \\
& BMR \cite{Bahnsen_2014_SIAM} & $.848 \pm .04$ & $.819 \pm .06$ & $.704 \pm .07$ \\
& Thresh \cite{Sheng_2006_AAAI} & $.837 \pm .05$ & $.829 \pm .03$ & $.703 \pm .06$ \\
& Our Method & $.880 \pm .02$ & $\textbf{.834} \pm .05$ & $\textbf{.742} \pm .06$ \\
\midrule
\multirow{6}{*}{\rotatebox{90}{MIMIC ICU}}
& Baseline & $.841 \pm .01$ & $.429 \pm .03$ & $.212 \pm .02$ \\
& SMOTE \cite{Chawla_2002_JAIR} & $.822 \pm .01$ & $.431 \pm .02$ & $.199 \pm .01$ \\
& BW & $.839 \pm .01$ & $.410 \pm .02$ & $.195 \pm .02$ \\
& BMR \cite{Bahnsen_2014_SIAM} & $\textbf{.852} \pm .01$ & $.405 \pm .04$ & $.202 \pm .02$ \\
& Thresh \cite{Sheng_2006_AAAI} & $.844 \pm .01$ & $.421 \pm .04$ & $.208 \pm .02$ \\
& Our Method & $.802 \pm .02$ & $\textbf{.477} \pm .04$ & $\textbf{.221} \pm .02$ \\
\bottomrule
\end{tabular}

\end{center}
\end{table}


\section{Discussion and Conclusion}
In this paper we propose a novel and theoretically-principled method of tackling class imbalance by tailoring pre-trained classifiers using confidence bounds. Importantly, our approach makes fewer assumptions on the underlying data distribution than alternative resampling methods, whilst not being specific to a single classifier -- like many of the cost-sensitive learning algorithms mentioned in Section~\ref{sec:Lit}.

Empirically, our method performs well in situations where there is a high level of class imbalance. This could act as a guidance to practitioners in order to decide which technique to use for the task at hand. However, it is important to highlight that it is not always feasible to find a solution to the optimisation problem proposed here. This is true for cases that involve pre-trained classifiers that have failed to learn a reasonable partition of the data in the projected space. In these situations, it is recommended to retrain the classifier from scratch with a cost-sensitive learning algorithm (where it would still be possible to set the bias term using our method) or if that is not an option, to use other model adjustment methods such as Thresholding or Bayes Minimum Risk.

\subsection{Limitations}
As mentioned above, our method relies on the pre-trained classifier having learnt a projected space for the training data which separates the classes well. This means that we do not provide a solution which is applicable to all classification scenarios. Practitioners are therefore required to have access to certain insights about the projected space learnt by their classifier or, alternatively, use a complex enough architecture or learning algorithm, such as \cite{Elsayed_2018_NeurIPS}, that guarantees that the representation learning stage is completed successfully. 
Cao et~al. state this is achievable by a neural network which has a sufficient number of parameters to fit to the training data \cite{Cao_2019_NuerIPS}.

The method we propose for relaxing the constraints by using slack variables is adequate to enforce a solution which aligns with the desired outcome, but it is not the only possibility and it does not converge if the training data classes in the projected space overlap significantly. Future work will explore this further to compare different alternatives, such as revisiting the original bounds and concentration inequalities in this scenario.

Finally, only binary data and classifiers are considered. This means that for multi-class data, a one-versus-rest or one-versus-one strategy must be taken. Ideally, an extension to our method to inherently deal with more than one class will need to be formulated to tackle multi-class settings in a more natural manner.

\begin{ack}
We would like to thank Jeff Clark and Chris McWilliams for their help with the evaluation on the MIMIC III dataset. 
This work is supported by the
UKRI Centre for Doctoral Training in Interactive AI EP/S022937/1, 
UKRI Turing AI Fellowship EP/V024817/1 
and the EU H2020 TAILOR, GA No 952215.
\end{ack}

\pagebreak
\bibliography{citations}

\pagebreak
\appendix
\section{Training Data Bound}
\label{appendix:training_data_bound}

Here we elaborate on how to derive Eq. \ref{eq:iid_error}. First, we combine the training dataset $X_t = \{x_1,..., x_N\}$ and test point, $x_{N+1}$, into one set $X = \{x_1,..., x_N, x_{N+1}\}$, where $|X| = N + 1$. 

Using the distance $d_n$ to order the set $X$, there are $(N+1)!$ possible combinations of ordering $X$. Out of these,
there are $N!$ possible combinations when the distance to the test point distance, $d_{N+1}$, is greater than all other points in the set. Therefore we can define the probability of this occurring as

\begin{equation*}
    \frac{N!}{(N+1)!} = \frac{N!}{N!(N+1)} = \frac{1}{(N+1)}.
\end{equation*}

\section{Continuous Slack Variables}
\label{appendix:cont_slacks}

An alternative to the binary slack variables, $\xi_n$, presented in Sec. \ref{sec:slacks} is to use continuous values such that $\xi_n = 0$ for $x_i$ that contribute towards $\hat{\bar{R}}_i$ and $N_i$, and $\xi_n = || x_n - \bar{\phi}_{S_i} ||$ for $x_i$ which are allowed to exist outside of the supports of the data and do not contribute towards $\hat{\bar{R}}_i$ and $N_i$. This definition of slack variables is more similar to how they are defined in soft-margin SVMs \cite{cortes1995support}, but adds additional complexity when compared to their binary counterpart.

Algorithm \ref{alg:ours} remains unchanged, apart from the new definition of the new continuous slack variables. Our code provides the option for a user to choose either binary or continuous slack variables \cite{clifford2024_code}.

Table \ref{tab:all_datasets_cont_slacks} shows the results on the datasets from Tab. \ref{tab:datasets} using continuous slack variables. The scores remain similar to those with the binary slack variables shown in Tab. \ref{tab:all_datasets} across all datasets.

\begin{table}[h]
\caption{Evaluation of all medical datasets with continuous slacks.}
\begin{center}
\label{tab:all_datasets_cont_slacks}

\begin{tabular}{@{}llccc@{}}
\toprule
& Methods & Accuracy & G-Mean & F1 \\
\midrule
\multirow{6}{*}{\rotatebox{90}{Pima Diabetes}}
& Baseline & $.540 \pm .04$ & $.181 \pm .23$ & $.137 \pm .18$ \\
& SMOTE \cite{Chawla_2002_JAIR} & $.595 \pm .01$ & $.495 \pm .03$ & $.396 \pm .04$ \\
& BW & $.585 \pm .02$ & $.476 \pm .06$ & $.373 \pm .08$ \\
& BMR \cite{Bahnsen_2014_SIAM} & $.661 \pm .02$ & $.628 \pm .04$ & $.575 \pm .05$ \\
& Thresh \cite{Sheng_2006_AAAI} & $.661 \pm .02$ & $.628 \pm .04$ & $.575 \pm .05$ \\
& Our Method & $\textbf{.672} \pm .03$ & $\textbf{.651} \pm .05$ & $\textbf{.611} \pm .08$ \\
\midrule
\multirow{6}{*}{\rotatebox{90}{Breast Cancer}}
& Baseline & $.917 \pm .02$ & $.917 \pm .02$ & $.913 \pm .02$ \\
& SMOTE \cite{Chawla_2002_JAIR} & $.793 \pm .09$ & $.768 \pm .12$ & $.737 \pm .14$ \\
& BW & $.921 \pm .02$ & $.921 \pm .02$ & $.918 \pm .02$ \\
& BMR \cite{Bahnsen_2014_SIAM} & $.887 \pm .08$ & $.880 \pm .09$ & $.869 \pm .11$ \\
& Thresh \cite{Sheng_2006_AAAI} & $.935 \pm .06$ & $.932 \pm .07$ & $.929 \pm .09$ \\
& Our Method & $\textbf{.943} \pm .02$ & $\textbf{.941} \pm .02$ & $\textbf{.945} \pm .02$ \\
\midrule
\multirow{6}{*}{\rotatebox{90}{Hepatitis}}
& Baseline & $\textbf{.827} \pm .06$ & $.623 \pm .15$ & $.485 \pm .16$ \\
& SMOTE \cite{Chawla_2002_JAIR} & $.804 \pm .02$ & $.299 \pm .31$ & $.224 \pm .24$ \\
& BW & $.768 \pm .09$ & $.681 \pm .09$ & $.499 \pm .10$ \\
& BMR \cite{Bahnsen_2014_SIAM} & $.804 \pm .05$ & $.656 \pm .11$ & $.493 \pm .10$ \\
& Thresh \cite{Sheng_2006_AAAI} & $.786 \pm .08$ & $.686 \pm .09$ & $.516 \pm .09$ \\
& Our Method & $.775 \pm .09$ & $\textbf{.714} \pm .06$ & $\textbf{.535} \pm .08$ \\
\midrule
\multirow{6}{*}{\rotatebox{90}{Heart Disease}}
& Baseline & $\textbf{.885} \pm .02$ & $.784 \pm .04$ & $.718 \pm .05$ \\
& SMOTE \cite{Chawla_2002_JAIR} & $.844 \pm .03$ & $.700 \pm .07$ & $.602 \pm .09$ \\
& BW & $.842 \pm .02$ & $.781 \pm .06$ & $.664 \pm .06$ \\
& BMR \cite{Bahnsen_2014_SIAM} & $.848 \pm .04$ & $.819 \pm .06$ & $.704 \pm .07$ \\
& Thresh \cite{Sheng_2006_AAAI} & $.837 \pm .05$ & $.829 \pm .03$ & $.703 \pm .06$ \\
& Our Method & $.878 \pm .02$ & $\textbf{.835} \pm .05$ & $\textbf{.740} \pm .06$ \\
\midrule
\multirow{6}{*}{\rotatebox{90}{MIMIC ICU}}
& Baseline & $.841 \pm .01$ & $.429 \pm .03$ & $.212 \pm .02$ \\
& SMOTE \cite{Chawla_2002_JAIR} & $.822 \pm .01$ & $.431 \pm .02$ & $.199 \pm .01$ \\
& BW & $.839 \pm .01$ & $.410 \pm .02$ & $.195 \pm .02$ \\
& BMR \cite{Bahnsen_2014_SIAM} & $\textbf{.852} \pm .01$ & $.405 \pm .04$ & $.202 \pm .02$ \\
& Thresh \cite{Sheng_2006_AAAI} & $.844 \pm .01$ & $.421 \pm .04$ & $.208 \pm .02$ \\
& Our Method & $.803 \pm .02$ & $\textbf{.476} \pm .04$ & $\textbf{.220} \pm .02$ \\
\bottomrule
\end{tabular}

\end{center}
\end{table}

\end{document}